\def\year{2021}\relax
\documentclass[letterpaper]{article} 
\usepackage{aaai21}  
\usepackage{times}  
\usepackage{helvet} 
\usepackage{courier}  
\usepackage[hyphens]{url}  
\usepackage{graphicx} 
\usepackage{natbib}  
\usepackage{caption} 
\usepackage{amsmath,bm}
\usepackage{amsfonts,amssymb}
\usepackage{graphicx}
\usepackage{caption}
\usepackage{subfigure}
\usepackage{booktabs}
\usepackage{multirow}
\usepackage{color}
\usepackage{algorithm}
\usepackage{algorithmic}
\usepackage[switch]{{lineno}}
\usepackage{colortbl}
\definecolor{mygray}{gray}{.9}

\def\u{\bm{u}}
\def\l{\bm{l}}
\nocopyright{\gdef\copyright@on{}}
\title{Improving the Certified Robustness of Neural Networks \\ via Consistency Regularization}
\author{
Anonymous
}

\title{Improving the Certified Robustness of Neural Networks via Consistency Regularization}
\author {
        Mengting Xu,
        Tao Zhang, 
        Zhongnian Li, 
        Daoqiang Zhang \footnote{Corresponding author}
        \\
}
\affiliations {
     Nanjing University of Aeronautics and Astronautics \\
    \{xumengting, dqzhang\}@nuaa.edu.cn
}

\begin{document}
\maketitle

\begin{abstract}
A range of defense methods have been proposed to improve the robustness of neural networks on adversarial examples, among which provable defense methods have been demonstrated to be effective to train neural networks that are certifiably robust to the attacker.  
However, most of these provable defense methods treat all examples equally during training process, which ignore the inconsistent constraint of certified robustness between correctly classified (natural) and misclassified examples. In this paper, we explore this inconsistency caused by misclassified examples and add a novel consistency regularization term to make better use of the misclassified examples. Specifically, we identified that the certified robustness of network can be significantly improved if the constraint of certified robustness on misclassified examples and correctly classified examples is consistent. Motivated by this discovery, we design a new defense regularization term called \textit{Misclassification Aware Adversarial Regularization} (MAAR), which constrains the output probability distributions of all examples in the certified region of the misclassified example. Experimental results show that our proposed MAAR achieves the best certified robustness and comparable accuracy on CIFAR-10 and MNIST datasets in comparison with several state-of-the-art methods.

\end{abstract}

\section{Introduction}

\begin{figure}[t]
	\centering                                                          
    \includegraphics[width=1.0\columnwidth]{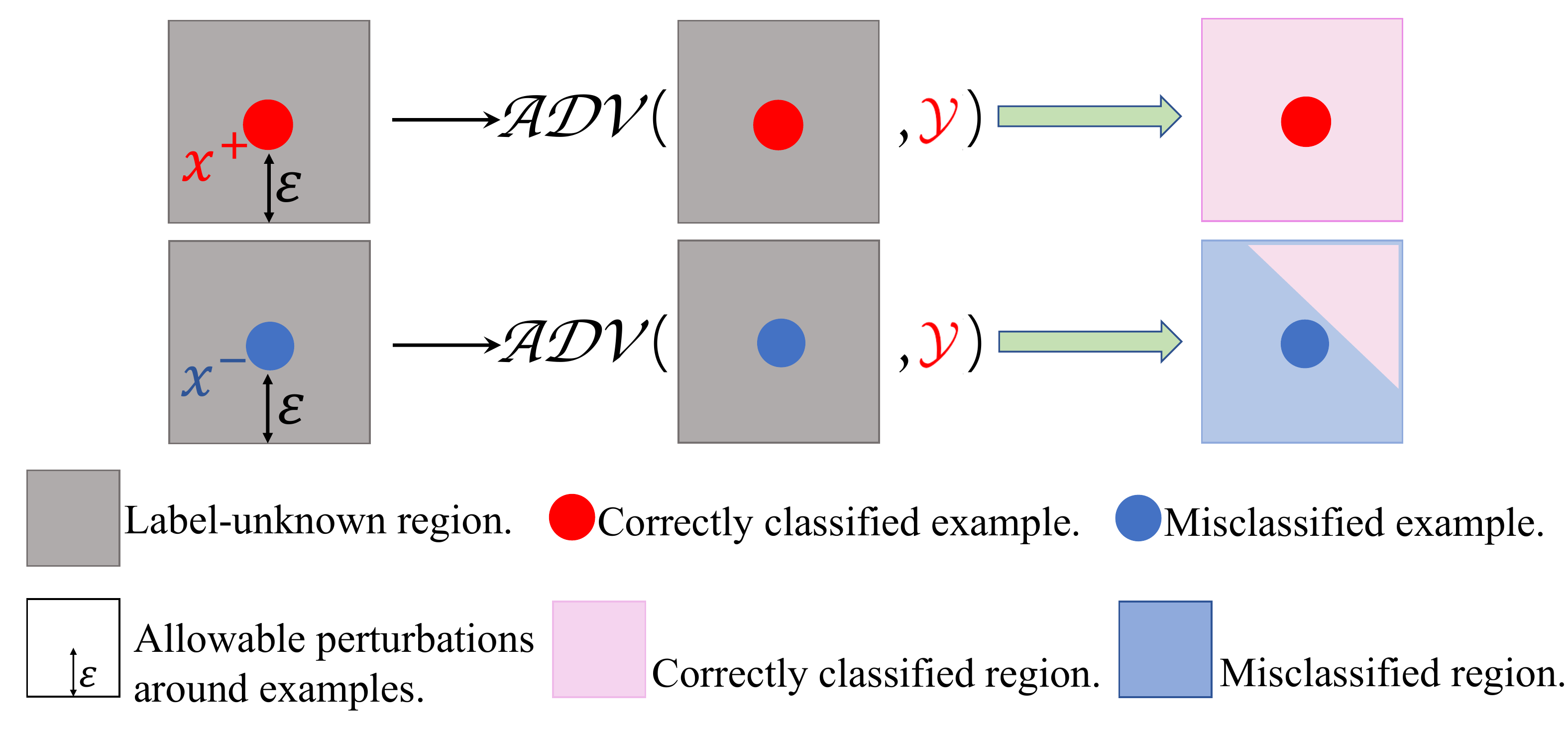}           
	\caption{Illustration about the inconsistent constraints of certified robustness between correctly classified and misclassified examples. 
		 As shown in the second row, $\mathcal{ADV}(\cdot, \mathcal{Y})$ loss on misclassified examples will keep the original accuracy instead of certified robustness while correctly classified examples keep the original accuracy as well as certified robustness as shown in the first row.
		This will lead the output probability distributions of all examples in the certified region of the misclassified examples to be unstable. 
}                      
	\label{fig:motivation}                                             
\end{figure}

\begin{figure}[t]
	\centering                                                          
	\includegraphics[width=1.0\columnwidth]{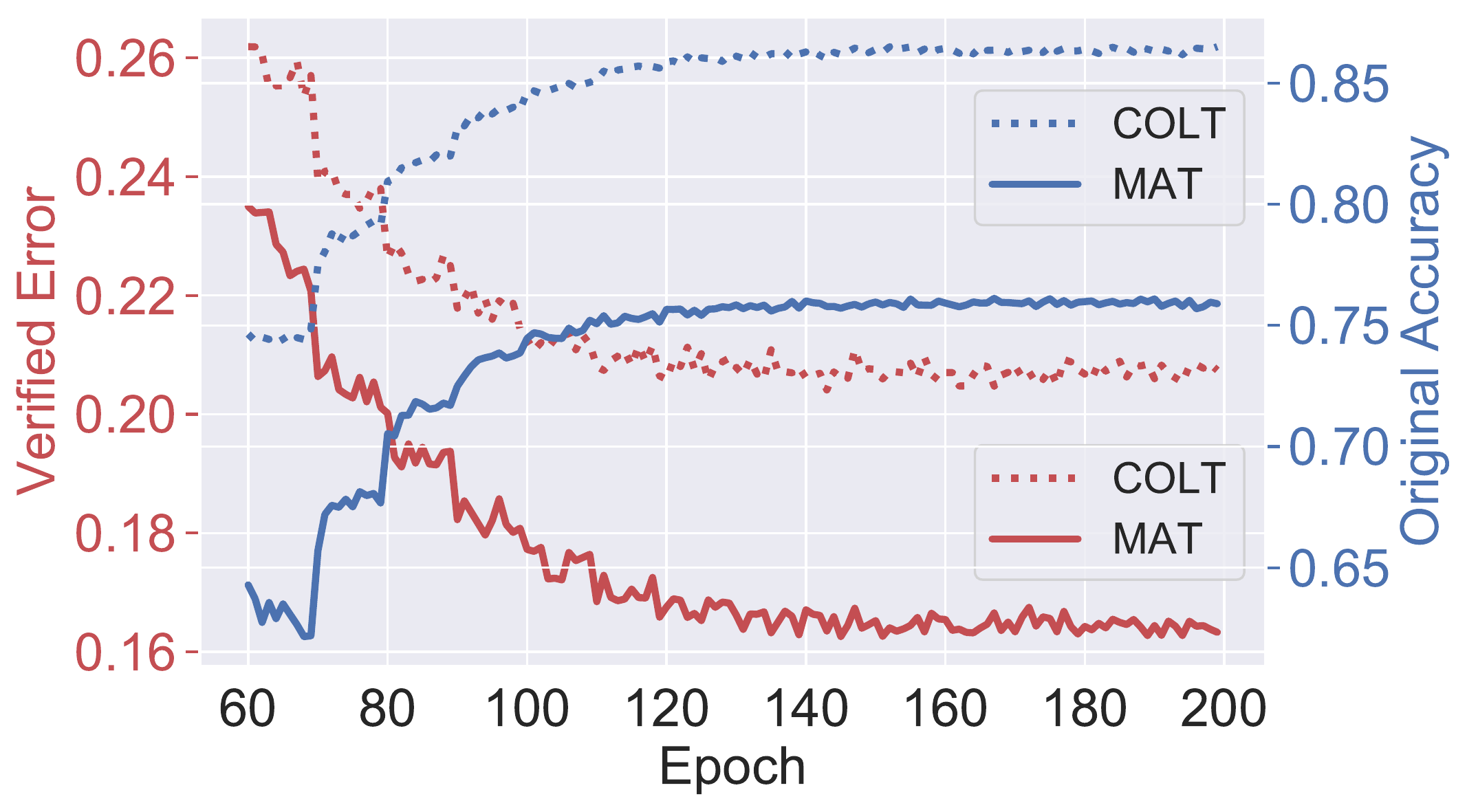}           
	\caption{Verified error (red lines) and original accuracy (blue lines) for COLT~\cite{balunovic2019adversarial} and Misclassification Aware Training (MAT). The dataset is CIFAR-10 with $L_\infty$ maximum perturbation $\epsilon = 2/255$. }                      
	\label{fig:motivation_exp}                                             
\end{figure}

\noindent Despite the widespread success of neural network on diverse tasks such as image classification~\cite{krizhevsky2012imagenet}, face and speech recognition~\cite{taigman2014deepface,hinton2012deep}.
Recent studies have highlighted the lack of robustness in state-of-the-art neural network models, e.g., a visually imperceptible adversarial image can be easily crafted to mislead a well-trained network~\cite{szegedy2013intriguing,goodfellow2014explaining}. 
The vulnerability to adversarial examples calls into question the safety-critical applications and services deployed by neural networks, including autonomous driving systems and malware detection protocols.

Considering the significance of adversarial robustness in neural network, a range of defense methods have been proposed. 
Adversarial training~\cite{goodfellow2014explaining,kurakin2016adversarial} which can be regarded as a data augmentation technique that trains neural networks on adversarial examples are highly robust against the strongest known adversarial attacks such as C\&W attack~\cite{carlini2017towards}, but it provides no guarantee --- it is unable to produce a certificate that there are no possible adversarial attack which could potentially break the model. To address this lack of guarantees, recent line of work on provable defense~\cite{wong2018provable,raghunathan2018certified,mirman2018differentiable,cohen2019certified} has proposed to train neural networks that no attacks within a certain region will alter the networks prediction. Moreover, recent work~\cite{balunovic2019adversarial} combines adversarial training and provable defense methods to train neural network with both high certified robustness and accuracy.

However, recall that the formal definition of certified robustness is conditioned on natural examples that are correctly classified~\cite{balunovic2019adversarial,wong2018provable}. Most provable defense methods treat equally in both correctly classified and misclassified examples during training process while evaluating certified robustness just on correctly classified examples. From this perspective, the effect of misclassified example on certified robustness is unknown. Therefore, it 
is not clear for the following questions: \textit{\textbf{(1) Do misclassified examples have effectiveness for improving certified robustness? (2) If yes, how can we make better use of misclassified examples to improve the certified robustness of neural network ? }}

In this paper, we investigate this significant aspect of certified robustness, and find that the misclassified examples do have excellent influence on the certified robustness. 
To validate this discovery, we conduct a proof-of-concept experiment on CIFAR-10~\cite{krizhevsky2009learning} with $L_\infty$ maximum perturbation $\epsilon=2/255$ based on convex layerwise adversarial training (COLT)~\cite{balunovic2019adversarial}. 

Different from COLT, which searches the potential adversarial examples on all natural examples within the convex relaxation region for adversarial training layerwisely, we dynamically select two subsets of natural training examples to investigate: 1) a subset of correctly classified examples $\mathcal{C}^+$ and 2) a subset of misclassified examples $\mathcal{C}^-$. Using these two subsets, we explore different loss functions to re-train the same network $h_{\bm{\theta}}$, and evaluate its original accuracy and verified error (The details of these metrics are in Appendix A.4) during the training process. Specifically, let $\mathcal{X}$ be natural input of the network with $\{\mathcal{C^+},\mathcal{C^-} \in \mathcal{X}\}$ and $\mathcal{Y}$ be its corresponding ground-truth label. $h_{\bm{\theta}}(\mathcal{X})$ is denoted as corresponding output label produced by the network. 
First, let's consider the traditional adversarial risk in COLT model: 
\begin{equation}
	COLT:  \mathcal{ADV}([\mathcal{C}^+]^\prime,\mathcal{Y})+\mathcal{ADV}([\mathcal{C}^-]^\prime,\mathcal{Y})
	\label{Equation1}
\end{equation}
where $\mathcal{ADV}(\cdot)$ is the adversarial training loss function, $[\mathcal{C}^+]^\prime$, $[\mathcal{C}^-]^\prime$ are adversarial examples produced by $\mathcal{C}^+$ and $\mathcal{C}^-$ within the perturbation sets, respectively.
As shown in the first part $\mathcal{ADV}([\mathcal{C}^+]^\prime,\mathcal{Y})$  in Equation~(\ref{Equation1}), the objectives of robustness constraint and original accuracy constraint are the same  for correctly classified examples:
\textit{Constraining the correctly classified examples $\mathcal{C^+}$ and all examples within their perturbations sets $[\mathcal{C}^+]^\prime$ to be close enough to the correct labels $\mathcal{Y}$.}
However, for misclassified examples, the constraints on robustness and accuracy in the
second part $\mathcal{ADV}([\mathcal{C}^-]^\prime,\mathcal{Y})$ are different:
\textit{Making the misclassified examples $\mathcal{C}^-$ and the examples in the perturbation set $[\mathcal{C}^-]^\prime$ close enough to the original labels $\mathcal{Y}$ will improve the original accuracy of the network, but it is undeniable that this will destroy the stability of the misclassified examples (the original label is the ``wrong label" for misclassified example), thereby reducing the certified robustness of the network.}
Based on the above observation, we find the constraints of certified robustness between correctly classified and misclassified examples are inconsistent, as shown in Figure~\ref{fig:motivation}.
To deal with this problem, we firstly propose to use the output label (the output label is the ``true label" for misclassified example) of the misclassified example during training process to keep the stability of misclassified examples. The adversarial risk is described as follows, which we call \textit{Misclassification Aware Training} (MAT):
\begin{equation}
MAT: \mathcal{ADV}([\mathcal{C}^+]^\prime,\mathcal{Y})+\mathcal{ADV}([\mathcal{C}^-]^\prime, h_{\bm{\theta}}(\mathcal{C}^-))
\end{equation}
where $h_{\bm{\theta}}(\mathcal{C}^-)$ denotes the output label corresponding to the input $\mathcal{C}^-$.

Interestingly, we find it that the misclassified examples have a significant effect on the final certified robustness of network (shown in Figure~\ref{fig:motivation_exp}). Compared with COLT (dashed red line), the verified error of MAT (red line) drops drastically. However, the original accuracy of MAT (blue line) is extremely lower than standard COLT (dashed blue line). 

In this paper, in order to make better use of misclassified examples, we propose a consistency regularization to constrain the output probability distributions of all examples in the certified region of the misclassified example.  The regularization term called Misclassification Aware Adversarial Regularization (MAAR) aims to encourage the output of  network to be stable against misclassified adversarial examples. In other words, MAAR focuses on solving the inconsistency of certified robustness on correctly classified and misclassified examples, which improves the final certified robustness of network. Meanwhile, MAAR does not change the training label during the training process, which alleviates the decrease of model accuracy.

Our main contributions are:
\begin{itemize}
\item We investigate the inconsistency on constraint of certified robustness caused by misclassified examples by a proof-of-concept experiment (i.e., \textit{Misclassification Aware Training} (MAT)). 
\item We propose a consistency regularization called \textit{Misclassification Aware Adversarial Regularization} (MAAR) which improves certified robustness by maintaining the stability of misclassified examples as well as relieving the degree of accuracy decline.
\item We show the effectiveness of MAAR by different networks and perturbations on two datasets. 
Specifically, MAAR achieves the state-of-the-art certified robustness of 62.8\% on CIFAR-10 with $2/255$ $L_\infty$ perturbations on 4-layer convolutional network as well as 97.3\%  on MNIST dataset with $L_\infty$ perturbation 0.1 on 3-layer convolutional network.
\end{itemize}

\section{Methods}

\subsection{Preliminaries}

\subsubsection{Base Classifier.}
For a $K$-class ($K\geq 2$) classification problem, denote a dataset $\{(\bm{x}_i ,y_i)\}_{i=1,\cdots,n}$ with distribution $\bm{x}_i \in \mathbb{R}^d $ as natural input and $y_i \in \{1,\cdots,K\}$ represents its corresponding true label, a classifier $h_{\bm{\theta}}$
with parameter $\bm{\theta}$ predicts the class of an input example $\bm{x}_i$:
\begin{equation}
h_{\bm{\theta}}(\bm{x}_i)=\mathop {\arg\max}_{k=1,\cdots,K} \bm{p}_k(\bm{x}_i,\bm{\theta}) 
\end{equation}
\begin{equation}
\bm{p}_k(\bm{x}_i,\bm{\theta})=\exp(\bm{z}_k(\bm{x}_i,\bm{\theta}))/ \sum_{k^\prime=1}^K \exp(\bm{z}_{k^\prime}(\bm{x}_i,\bm{\theta}))
\end{equation}
where $\bm{z}_k(\bm{x}_i,\bm{\theta})$ is the logits output of the network with respect to class $k$, and $\bm{p}_k(\bm{x}_i,\bm{\theta})$ is the probability (softmax on logits) of $\bm{x}_i$ belonging to class $k$.

\subsubsection{Adversarial Risk.} 
The adversarial risk~\cite{madry2017towards} on dataset $\{(\bm{x}_i ,y_i)\}_{i=1,\cdots,n}$ and classifier $h_{\bm{\theta}}$ output probability $\bm{p}(\bm{x})$ can be defined as follows:
\begin{equation}
\mathcal{ADV}(\bm{p}(\bm{x}^\prime),y)=\frac{1}{n} \sum_{i=1}^{n} \max_{\bm{x}_i^\prime \in \mathcal{B}_\epsilon(\bm{x}_i)} \mathcal{L} (\bm{p}(\bm{x}_i^\prime),y_i)
\end{equation}
where $\mathcal{L}$ is the loss function such as commonly used cross entropy loss, and $\mathcal{B}_\epsilon(\bm{x}_i)=\{\bm{x}:||\bm{x}-\bm{x}_i||_p\leq\epsilon\}$ denotes the $L_p$-norm ball centered at $\bm{x}_i$ with radius $\epsilon$. We will focus on the $L_\infty$-ball in this paper.

\subsubsection{Original Training Risk in COLT.}
The original training risk in COLT~\cite{balunovic2019adversarial} is defined as follows:
\begin{equation}
\label{equ:colt}
\mathcal{R}_{COLT}(h_{\bm{\theta}},\bm{x}_i) :=\mathcal{L} _{ori}(\bm{p}(\bm{x}_i),y_i)+\mathcal{ADV}(\bm{p}(\bm{x}_i^\prime),y_i)
\end{equation}
where $\mathcal{L}_{ori}(\cdot)$ is the original training loss function such as cross entropy loss and $\bm{x}_i^\prime \in \mathcal{B}_\epsilon(\bm{x}_i)$. 

\subsection{Misclassification Aware Adversarial Regularization}
To differentiate and explore the effect of misclassified examples, we reformulate the training risk based on the prediction of the current network $h_{\bm{\theta}}$. 
Specifically, we split the natural training examples into two subset according to $h_{\bm{\theta}}$, with one subset of correctly classified examples ($\mathcal{C}_{h_{\bm{\theta}}}^+$) and one subset of misclassified examples ($\mathcal{C}_{h_{\bm{\theta}}}^-$):
\begin{align}
\mathcal{C}_{h_{\bm{\theta}}}^+=\{i:i\in[n], h_{\bm{\theta}}(\bm{x}_i)=y_i\}\\
\mathcal{C}_{h_{\bm{\theta}}}^-=\{i:i\in[n], h_{\bm{\theta}}(\bm{x}_i)\neq y_i\}
\end{align}

With the purpose of avoiding excessive reduction of the original accuracy as well as keeping the consistency of certified robustness on two subsets, we regularize misclassified examples by an additional term (a KL-divergence term that was used previously in~\cite{wang2019improving,zhang2019theoretically,zheng2016improving}) rather than changing the training labels.
 The proposed regularization term aims to constrain the output probability distributions of all examples in the certified region of the misclassified example, thus improving the certified robustness of network.
The improved training risk of misclassified examples is formulated as follows:

\begin{equation}
\label{equ:R-}
\begin{split}
\mathcal{R}^-(h_{\bm{\theta}}, \bm{x}_i) &:=  \mathcal{L} _{ori}(\bm{p}(\bm{x}_i),y_i)+\mathcal{ADV}(\bm{p}(\bm{x}_i^\prime),y_i)\\
&+\mathcal{KL}(\bm{p}(\bm{x}_i)||\bm{p}(\bm{x}_i^\prime))
\end{split}
\end{equation}
where
\begin{equation}
\mathcal{KL}(\bm{p}(\bm{x}_i)||\bm{p}(\bm{x}_i^\prime))=\sum_{k=1}^K \bm{p}_k(\bm{x}_i,\bm{\theta})\log \frac{\bm{p}_k(\bm{x}_i,\bm{\theta})}{\bm{p}_k(\bm{x}_i^\prime,\bm{\theta})}
\end{equation}
measures the difference of two distributions.

For correctly classified examples, we simply use original training risk, i.e.,
\begin{equation}
\label{equ:R+}
\begin{split}
\mathcal{R}^+(h_{\bm{\theta}}, \bm{x}_i) &:=  \mathcal{L} _{ori}(\bm{p}(\bm{x}_i),y_i)+\mathcal{ADV}(\bm{p}(\bm{x}_i^\prime),y_i)
\end{split}
\end{equation}

Finally, by combining the two training risk terms (i.e., Equation~(\ref{equ:R-}) and Equation~(\ref{equ:R+})), we train a network that minimizes the following risk:
\begin{equation}
\label{equ:re}
\begin{split}
\mathcal{R}(h_{\bm{\theta}},\bm{x})&:=\frac{1}{n}(\sum_{\bm{x}_i \in \mathcal{C}_{h_{\bm{\theta}}}^+} \mathcal{R}^+(h_{\bm{\theta}},\bm{x}_i) + \sum_{\bm{x}_i \in \mathcal{C}_{h_{\bm{\theta}}}^-} \mathcal{R}^-(h_{\bm{\theta}},\bm{x}_i) )\\
& =\frac{1}{n} \sum_{i=1}^{n} \{\mathcal{L} _{ori}(\bm{p}(\bm{x}_i),y_i)+\mathcal{ADV}(\bm{p}(\bm{x}_i^\prime),y_i)\\
& +\mathcal{KL}(\bm{p}(\bm{x}_i)||\bm{p}(\bm{x}_i^\prime)) \cdot \mathbb{I}(h_{\bm{\theta}}(\bm{x}_i)\neq y_i)\}
\end{split}
\end{equation}
where $\mathbb{I}(h_{\bm{\theta}}(\bm{x}_i)\neq y_i)$ is the indicator function. $\mathbb{I}(h_{\bm{\theta}}(\bm{x}_i)\neq y_i) =1$ if $h_{\bm{\theta}}(\bm{x}_i)\neq y_i$, and $\mathbb{I}(h_{\bm{\theta}}(\bm{x}_i)\neq y_i) =0$ otherwise.

\begin{table*}
	\centering
	\caption{Comparison with the state-of-the-art methods. Accuracy and certified robustness evaluated with $L_\infty$ perturbation 2/255 and 8/255 on CIFAR-10 dataset, $L_\infty$ perturbation 0.1 on MNIST dataset. ACC: Accuracy, CR: Certified robustness. }
	\label{tab:result}
	\centering
	\begin{tabular}{c cccc c cc}
		\toprule
		\multirow{3}{*}{Method} & \multicolumn{4}{c}{\textbf{CIFAR-10}} & & \multicolumn{2}{c}{\textbf{MNIST} }\\
		\cmidrule{2-5} \cmidrule{7-8}
		& \multicolumn{2}{c}{$\epsilon=2/255$} & \multicolumn{2}{c}{$\epsilon=8/255$} & & \multicolumn{2}{c}{$\epsilon=0.1$} 
		\\
		\cmidrule{2-5} \cmidrule{7-8}
		& ACC(\%) & CR(\%) & ACC(\%) & CR(\%)& & ACC(\%) & CR(\%)
		\\
		\midrule
		Our work(MAAR) & 77.7 & \textbf{62.8} & 47.6 & \textbf{29.8} & & 99.1 & \textbf{97.3} \\
		COLT~\shortcite{balunovic2019adversarial} &80.0 & 58.6 & 51.3 & 26.7 & & 99.2 & 97.1 \\
		CROWN-IBP~\shortcite{zhang2019towards} & 61.6 & 48.6 & 48.5& 26.3& & 98.7& 96.6 \\
		IBP~\shortcite{gowal2018effectiveness} &58.0& 47.8&47.8&24.9& & 98.8& 95.8 \\
		Xiao et al.\shortcite{xiao2018training} & 61.1 & 45.9 & 40.5 & 20.3 & & 99.0 & 95.6 \\
		Mirman et al.\shortcite{mirman2019provable} & 62.3 & 45.5 & 46.2 & 27.2 & & 98.7 & 96.8 \\
		\bottomrule		
	\end{tabular}
\end{table*}

\subsubsection{Optimization for Regularization Term.}
As presented in Equation~(\ref{equ:re}), the new training risk is a regularized adversarial risk with regularization term $\frac{1}{n}\sum_{i=1}^{n} \{\mathcal{KL}(\bm{p}(\bm{x}_i)||\bm{p}(\bm{x}_i^\prime)) \cdot \mathbb{I}(h_{\bm{\theta}}(\bm{x}_i)\neq y_i)\}$. However, the indicator function cannot be directly optimized if we conduct a hard decision during the training process. In this study, we propose to use a soft decision scheme by replacing $\mathbb{I}(h_{\bm{\theta}}(\bm{x}_i)\neq y_i)$ with the output probability $1-\bm{p}_{y_i}(\bm{x}_i,\bm{\theta})$. 
The output probability will be large for misclassified examples and small for correctly classified examples, by which we could provide a approximate solution for 0-1 optimization problem .


\subsubsection{The Overall Objective.}
Based on the regularization optimization, the   objective function of our proposed \textit{Misclassification Aware Adversarial Regularization} (MAAR) is formulated as:
\begin{equation}
\mathcal{R}^{MAAR}(\bm{\theta})=\frac{1}{n} \sum_{i=1}^{n} \mathcal{L}^{MAAR}(\bm{x}_i,y_i,\bm{\theta})
\end{equation}
where $\mathcal{L}^{MAAR}(\bm{x}_i,y_i,\bm{\theta})$ is defined as:
\begin{equation}
\label{equ:obj}
\begin{split}
\mathcal{L}^{MAAR}(&\bm{x}_i, y_i,\bm{\theta}) = \mathcal{L} _{ori}(\bm{p}(\bm{x}_i),y_i)+\mathcal{ADV}(\bm{p}(\bm{x}_i^\prime),y_i)\\
&+\lambda \cdot \mathcal{KL}(\bm{p}(\bm{x}_i)||\bm{p}(\bm{x}_i^\prime)) \cdot (1-\bm{p}_{y_i}(\bm{x}_i,\bm{\theta}))
\end{split}
\end{equation}
Here, $\lambda$ are tunable scaling paremeters and fixed for all training examples. 
The sensitivity of $\lambda$ is described in Appendix C.1.
For more details on the training procedure, see Appendix A.

\section{Results}

\begin{figure}[h]
	\centering
	\includegraphics[width=0.9\columnwidth]{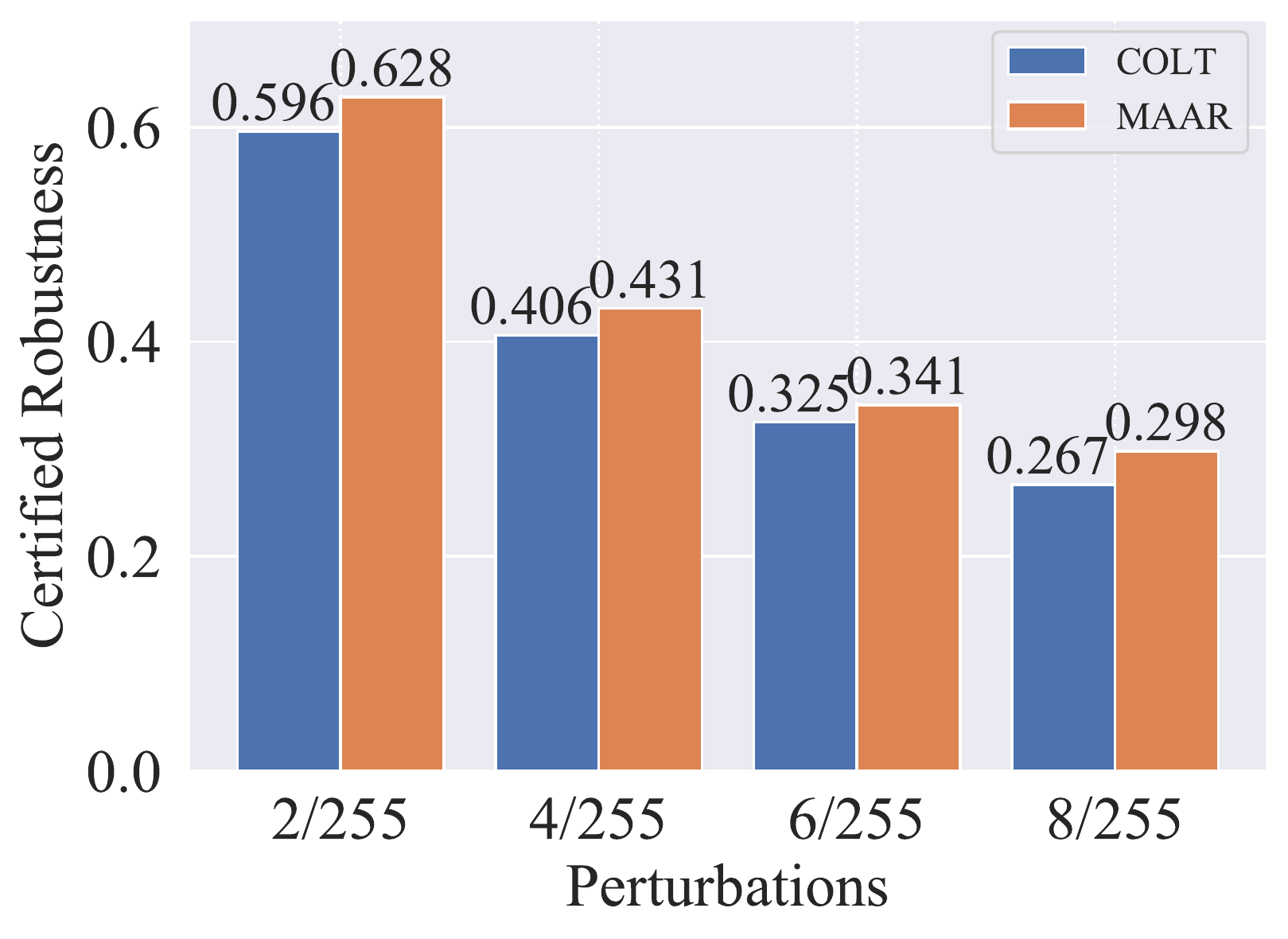} 
	\caption{Certified robustness of COLT and MAAR with different perturbations.}
	\label{fig:diffper}
\end{figure}

\subsection{Certification under Different Perturbations.}
We evaluate the effectiveness of  MAAR on certified robustness with different perturbations. As shown in Figure~\ref{fig:diffper}, the certified robustness of MAAR (orange bar) is obviously higher than COLT (blue bar) during different perturbations. 
Specifically, MAAR achieves the state-of-the-art certified robustness (i.e., 62.8\%) compared with COLT (59.6\%) when $\epsilon = 2/255$.
We also show the effectiveness of MAAR during laywise training process in Appendix C.2.
What's more, further experimental results show that our MAAR achieves better certified robustness (62.8\%) when comparision with giving all examples the same weight (60.8\%).

\subsection{Comparison with Prior Work.}
We compare our MAAR with COLT~\cite{balunovic2019adversarial}, CROWN-IBP~\cite{zhang2019towards}, IBP~\cite{gowal2018effectiveness} in the same network architecture, parameter settings, and certification progress. We also list the results reported in literature of Xiao et al.\shortcite{xiao2018training} and Mirman et al.\shortcite{mirman2019provable} in Table~\ref{tab:result}. The network sizes and training parameters on two datasets (i.e., CIFAR-10 and MNIST) can be found in Appendix B.
\subsubsection{CIFAR-10.}
For the $L_\infty$ perturbation 2/255. Experiment results show that MAAR substantially outperforms its competitors by certified robustness (i.e., 62.8\%). 
Besides, the accuracy of our method also outperforms other works except COLT. 
This is because one side-effect of our regularization is that it will maintain the distribution around misclassified examples, which will decrease the accuracy in comparison with COLT. 
Actually, the accuracy–robustness trade-off has been proved to exist in predictive models when training robust models~\cite{tsipras2018robustness,zhang2019theoretically}.

We also run the same experiment for $L_\infty $ perturbation 8/255. When under the same experimental settings, MAAR also achieves the best certified robustness (i.e., 29.8\%). 
Here we do not compare with the certified robustness (i.e., 32,0\%) reported in literature of IBP~\cite{gowal2018effectiveness} as their results were found to be not reproducible~\cite{mirman2019provable,zhang2019towards}.
\subsubsection{MNIST.}
To futher evaluate the effectiveness of our method, we also conduct experiments on MNIST dataset with $L_\infty$ perturbation 0.1. 
We report the full results in Table~\ref{tab:result}, MAAR also achieve the state-of-the-art certified robustness (i.e., 97.3\%) comparable with best results from prior work (i.e., 97.1\%). 
\section{Conclusion }
In this paper, we investigated the inconsistent constraint of certified robustness between correctly classified and misclassified and find that misclassified examples have a recognizable impact on the final certified robustness of network. Based on this observation, we designed a consistency regularization term which constrains the output probability distributions of examples in the certified region of the misclassified example. Our method, named Misclassification Aware Adversarial Regularization (MAAR), achieves the state-of-the-art certified robustness of 62.8\%  on CIFAR-10 with 2/255 $L_\infty$ perturbation as well as 97.3\% on MNIST dataset with $L_\infty$ perturbation 0.1. The method is general and can be instantiated with most of training risk. 

In the future, we plan to investigate the association between accuracy and certified robustness among different neural networks, and apply our method to more provable defense frameworks.

\bibliography{refer}

\begin{thebibliography}{24}
\providecommand{\natexlab}[1]{#1}
\providecommand{\url}[1]{\texttt{#1}}
\providecommand{\urlprefix}{URL }
\expandafter\ifx\csname urlstyle\endcsname\relax
  \providecommand{\doi}[1]{doi:\discretionary{}{}{}#1}\else
  \providecommand{\doi}{doi:\discretionary{}{}{}\begingroup
  \urlstyle{rm}\Url}\fi

\bibitem[{Balunovic and Vechev(2020)}]{balunovic2019adversarial}
Balunovic, M.; and Vechev, M. 2020.
\newblock Adversarial training and provable defenses: Bridging the gap.
\newblock In \emph{International Conference on Learning Representations}.

\bibitem[{Carlini and Wagner(2017)}]{carlini2017towards}
Carlini, N.; and Wagner, D. 2017.
\newblock Towards evaluating the robustness of neural networks.
\newblock In \emph{IEEE symposium on security and privacy (SP)}, 39--57. IEEE.

\bibitem[{Cohen, Rosenfeld, and Kolter(2019)}]{cohen2019certified}
Cohen, J.; Rosenfeld, E.; and Kolter, Z. 2019.
\newblock Certified Adversarial Robustness via Randomized Smoothing.
\newblock In \emph{International Conference on Machine Learning}, 1310--1320.

\bibitem[{Ehlers(2017)}]{ehlers2017formal}
Ehlers, R. 2017.
\newblock Formal verification of piece-wise linear feed-forward neural
  networks.
\newblock In \emph{International Symposium on Automated Technology for
  Verification and Analysis}, 269--286. Springer.

\bibitem[{Goodfellow, Shlens, and Szegedy(2015)}]{goodfellow2014explaining}
Goodfellow, I.~J.; Shlens, J.; and Szegedy, C. 2015.
\newblock Explaining and harnessing adversarial examples.
\newblock In \emph{International Conference on Learning Representations}.

\bibitem[{Gowal et~al.(2018)Gowal, Dvijotham, Stanforth, Bunel, Qin, Uesato,
  Arandjelovic, Mann, and Kohli}]{gowal2018effectiveness}
Gowal, S.; Dvijotham, K.; Stanforth, R.; Bunel, R.; Qin, C.; Uesato, J.;
  Arandjelovic, R.; Mann, T.; and Kohli, P. 2018.
\newblock On the effectiveness of interval bound propagation for training
  verifiably robust models.
\newblock \emph{arXiv preprint arXiv:1810.12715} .

\bibitem[{Hinton et~al.(2012)Hinton, Deng, Yu, Dahl, Mohamed, Jaitly, Senior,
  Vanhoucke, Nguyen, Sainath et~al.}]{hinton2012deep}
Hinton, G.; Deng, L.; Yu, D.; Dahl, G.~E.; Mohamed, A.-r.; Jaitly, N.; Senior,
  A.; Vanhoucke, V.; Nguyen, P.; Sainath, T.~N.; et~al. 2012.
\newblock Deep neural networks for acoustic modeling in speech recognition: The
  shared views of four research groups.
\newblock \emph{IEEE Signal processing magazine} 29(6): 82--97.

\bibitem[{Kingma and Ba(2014)}]{kingma2014adam}
Kingma, D.~P.; and Ba, J. 2014.
\newblock Adam: A method for stochastic optimization.
\newblock \emph{arXiv preprint arXiv:1412.6980} .

\bibitem[{Krizhevsky, Hinton et~al.(2009)}]{krizhevsky2009learning}
Krizhevsky, A.; Hinton, G.; et~al. 2009.
\newblock Learning multiple layers of features from tiny images .

\bibitem[{Krizhevsky, Sutskever, and Hinton(2012)}]{krizhevsky2012imagenet}
Krizhevsky, A.; Sutskever, I.; and Hinton, G.~E. 2012.
\newblock Imagenet classification with deep convolutional neural networks.
\newblock In \emph{Advances in Neural Information Processing Systems},
  1097--1105.

\bibitem[{Kurakin, Goodfellow, and Bengio(2016)}]{kurakin2016adversarial}
Kurakin, A.; Goodfellow, I.; and Bengio, S. 2016.
\newblock Adversarial examples in the physical world.
\newblock \emph{arXiv preprint arXiv:1607.02533} .

\bibitem[{Madry et~al.(2017)Madry, Makelov, Schmidt, Tsipras, and
  Vladu}]{madry2017towards}
Madry, A.; Makelov, A.; Schmidt, L.; Tsipras, D.; and Vladu, A. 2017.
\newblock Towards deep learning models resistant to adversarial attacks.
\newblock \emph{arXiv preprint arXiv:1706.06083} .

\bibitem[{Mirman, Gehr, and Vechev(2018)}]{mirman2018differentiable}
Mirman, M.; Gehr, T.; and Vechev, M. 2018.
\newblock Differentiable abstract interpretation for provably robust neural
  networks.
\newblock In \emph{International Conference on Machine Learning}, 3578--3586.

\bibitem[{Mirman, Singh, and Vechev(2019)}]{mirman2019provable}
Mirman, M.; Singh, G.; and Vechev, M. 2019.
\newblock A provable defense for deep residual networks.
\newblock \emph{arXiv preprint arXiv:1903.12519} .

\bibitem[{Raghunathan, Steinhardt, and Liang(2018)}]{raghunathan2018certified}
Raghunathan, A.; Steinhardt, J.; and Liang, P. 2018.
\newblock Certified Defenses against Adversarial Examples.
\newblock In \emph{International Conference on Learning Representations}.

\bibitem[{Szegedy et~al.(2014)Szegedy, Zaremba, Sutskever, Bruna, Erhan,
  Goodfellow, and Fergus}]{szegedy2013intriguing}
Szegedy, C.; Zaremba, W.; Sutskever, I.; Bruna, J.; Erhan, D.; Goodfellow, I.;
  and Fergus, R. 2014.
\newblock Intriguing properties of neural networks.
\newblock In \emph{International Conference on Learning Representations}.

\bibitem[{Taigman et~al.(2014)Taigman, Yang, Ranzato, and
  Wolf}]{taigman2014deepface}
Taigman, Y.; Yang, M.; Ranzato, M.; and Wolf, L. 2014.
\newblock Deepface: Closing the gap to human-level performance in face
  verification.
\newblock In \emph{Proceedings of the IEEE conference on computer vision and
  pattern recognition}, 1701--1708.

\bibitem[{Tsipras et~al.(2018)Tsipras, Santurkar, Engstrom, Turner, and
  Madry}]{tsipras2018robustness}
Tsipras, D.; Santurkar, S.; Engstrom, L.; Turner, A.; and Madry, A. 2018.
\newblock Robustness may be at odds with accuracy.
\newblock \emph{arXiv preprint arXiv:1805.12152} .

\bibitem[{Wang et~al.(2019)Wang, Zou, Yi, Bailey, Ma, and
  Gu}]{wang2019improving}
Wang, Y.; Zou, D.; Yi, J.; Bailey, J.; Ma, X.; and Gu, Q. 2019.
\newblock Improving adversarial robustness requires revisiting misclassified
  examples.
\newblock In \emph{International Conference on Learning Representations}.

\bibitem[{Wong and Kolter(2018)}]{wong2018provable}
Wong, E.; and Kolter, Z. 2018.
\newblock Provable defenses against adversarial examples via the convex outer
  adversarial polytope.
\newblock In \emph{International Conference on Machine Learning}, 5286--5295.

\bibitem[{Xiao et~al.(2018)Xiao, Tjeng, Shafiullah, and
  Madry}]{xiao2018training}
Xiao, K.~Y.; Tjeng, V.; Shafiullah, N. M.~M.; and Madry, A. 2018.
\newblock Training for Faster Adversarial Robustness Verification via Inducing
  ReLU Stability.
\newblock In \emph{International Conference on Learning Representations}.

\bibitem[{Zhang et~al.(2019{\natexlab{a}})Zhang, Chen, Xiao, Gowal, Stanforth,
  Li, Boning, and Hsieh}]{zhang2019towards}
Zhang, H.; Chen, H.; Xiao, C.; Gowal, S.; Stanforth, R.; Li, B.; Boning, D.;
  and Hsieh, C.-J. 2019{\natexlab{a}}.
\newblock Towards stable and efficient training of verifiably robust neural
  networks.
\newblock \emph{arXiv preprint arXiv:1906.06316} .

\bibitem[{Zhang et~al.(2019{\natexlab{b}})Zhang, Yu, Jiao, Xing, Ghaoui, and
  Jordan}]{zhang2019theoretically}
Zhang, H.; Yu, Y.; Jiao, J.; Xing, E.~P.; Ghaoui, L.~E.; and Jordan, M.~I.
  2019{\natexlab{b}}.
\newblock Theoretically principled trade-off between robustness and accuracy.
\newblock \emph{arXiv preprint arXiv:1901.08573} .

\bibitem[{Zheng et~al.(2016)Zheng, Song, Leung, and
  Goodfellow}]{zheng2016improving}
Zheng, S.; Song, Y.; Leung, T.; and Goodfellow, I. 2016.
\newblock Improving the robustness of deep neural networks via stability
  training.
\newblock In \emph{Proceedings of the IEEE conference on computer vision and
  pattern recognition}, 4480--4488.

\end{thebibliography}

\clearpage

\onecolumn
\section{Appendix A. Algorithmic Details}
\subsection{A.1 MAAR Algorithmic}

\begin{algorithm} [h]
	\caption{Misclassification Aware Adversarial Regularization (MAAR)}  
	\label{alg1}
	\begin{algorithmic}
		\STATE {\bfseries Input} {
			$d$-layer network $h_{\bm{\theta}}$, training set $(\mathcal{X},\mathcal{Y})$, learning rate $\eta$, step size $\alpha$, inner steps $n$, tunable scaling parameters $\lambda$, perturbation $\epsilon$.
		}
		\FOR{$l\leq d$}
		\FOR{$j \leq n_{epochs}$}
		\STATE Sample mini-batch: \STATE$(\bm{x}_1,y_1),(\bm{x}_2,y_2)\cdots,(\bm{x}_b,y_b)\} \sim
		(\mathcal{X},\mathcal{Y});$
		\STATE Compute convex relaxations: $\mathbb{C}_l(\bm{x}_1),\cdots,\mathbb{C}_l(\bm{x}_b)$;
		\STATE Initialize: $\bm{x}_1^\prime \sim \mathbb{C}_l(\bm{x}_1), \cdots, \bm{x}_b^\prime \sim \mathbb{C}_l(\bm{x}_b)$;
		\FOR{$i \leq b$}
		\STATE Update in parallel $n$ times: 
		\STATE $\bm{x}_i^\prime \leftarrow \Pi_{\mathbb{C}_l(\bm{x}_i)}(\bm{x}_i^\prime + \alpha \nabla_{\bm{x}_i^\prime} \mathcal{ADV}(h_{\bm{\theta}}^{l+1:d}(\bm{x}_i^\prime),y_i))$;
		\ENDFOR
		\STATE $\mathcal{L}(h_{\bm{\theta}}^{l+1:d}(\bm{x}_i^\prime),y_i)\leftarrow  \mathcal{L} _{ori}(\bm{p}(\bm{x}_i),y_i) +\mathcal{ADV}(\bm{p}(\bm{x}_i^\prime),y_i)
		+\lambda \cdot \mathcal{KL}(\bm{p}(\bm{x}_i)||\bm{p}(\bm{x}_i^\prime)) \cdot (1-\bm{p}_{y_i}(\bm{x}_i,\bm{\theta}))$
		\STATE Update parameters: \STATE $\bm{\theta}\leftarrow\bm{\theta}-\eta\cdot\frac{1}{b}\sum_{i=1}^{b}\nabla_{\bm{\theta}}\mathcal{L}(h_{\bm{\theta}}^{l+1:d}(\bm{x}_i^\prime),y_i)$;
		\ENDFOR
		\STATE  Freeze parameters $\bm{\theta}_{l+1}$ of layer function $h_\theta^{l+1}$.
		\ENDFOR
		\STATE {\bfseries Output} {Certified robust neural network $h_{\bm{\theta}}$}
	\end{algorithmic}
\end{algorithm}

\subsection{A.2 Training}

The layerwise training \cite{balunovic2019adversarial} has been adopted in our MAAR's training. Compared with standard adversarial training in~\cite{madry2017towards}, layerwise training not only performs adversarial attack on the input domain, but also searches potential adversarial examples in the hidden layer of the convex perturbation region. It will freeze the previous layers and stop back-propagation after the update of the current layer. 

Each convex region of layer $i$ is represented as a set $\mathbb{C}_i(\bm{x}) =\{\bm{a}_i+\bm{A}_i\bm{e} | \bm{e} \in \left[ -1,1\right]^{m_i}\}$~\cite{wong2018provable}, vector $\bm{a}_i$ represents the center of the set and the matrix $\bm{A}_i$ represents the affine transformation of the hypercube $\left[-1,1\right]^{m_i}$. Propagation of $\bm{a}_i$, $\bm{A}_i$ through the network can be described as follows:

\begin{equation}
\bm{a}_{i+1} =
\begin{cases}
\bm{\Lambda}_{i+1} \bm{a}_{i},  & \textit{\#ReLU Layer}  \\
\bm{W}_{i+1} \bm{a}_{i}+\bm{b}_{i+1}, & \textit{\#Conv. \& FC. layer}  
\end{cases}
\quad
\quad
\mathrm{s.t.}
\begin{cases}
&\bm{a}_0:=\frac{1}{2}(\bm{x}^l+\bm{x}^u) \\
&\bm{\Lambda}_{i+1}:=\frac{\bm{u}_{i}}{\bm{u}_{i}-\bm{l}_{i}}
\end{cases}
\end{equation}

\begin{equation}
\label{equ:A}
\bm{A}_{i+1} =
\begin{cases}
\left[\bm{\Lambda}_{i+1} \bm{A}_{i} ,\bm{M}_{i+1} \right],  &\textit{\#ReLU Layer} \\
\bm{W}_{i+1} \bm{A}_{i}, & \textit{\#Conv. \& FC. layer} 
\end{cases}
\quad \quad 
\mathrm{s.t.}
\begin{cases}
& \bm{A}_0 := \frac{1}{2} \bm{I}_{m_0}(\bm{x}^u-\bm{x}^l) \\
& \bm{M}_{i+1}:=\frac{-\bm{u}_{i} \bm{l}_{i}}{2(\bm{u}_{i}-\bm{l}_{i})} 
\end{cases}
\end{equation}
where $\bm{W}_{i+1}$ and $\bm{b}_{i+1}$ are the weights matrix and bias matrix of layer $i+1$, respectively. $\bm{x}^l = \max(0, \bm{x}-\epsilon)$ and $\bm{x}^u = \min(1, \bm{x}+\epsilon)$, $\epsilon$ is the $L_\infty$ radius which we are certifying, $\left[ \cdot \right]$denotes concatenation of matrics along the column dimension.
Specifically, we first compute lower bound $\bm{l}_i$ and upper bound $\bm{u}_i$ in the set $\mathbb{C}_i(\bm{x})$:
\begin{equation}
	\l_i = \bm{a}_i -|\bm{A}_{i}| 
\end{equation}
\begin{equation}
	\u_i = \bm{a}_i + |\bm{A}_i|
\end{equation}
Then, we need to compute $||\bm{A}_i||_1$, which is the $L_1$ norm of $\bm{A}_i$. According to Equation~(\ref{equ:A}), we can get:
\begin{equation}
\bm{A}_i =\left[\bm{W}_i\Lambda_{i-1}\bm{W}_{i-2}\Lambda_{i-3}\cdots\bm{M}_0,\cdots,\bm{W}_i\Lambda_{i-1}\bm{W}_{i-2}\bm{M}_{i-3},\bm{W}_i\bm{M}_{i-1}\right]
\end{equation}
We use Cauchy random projections to efficiently estimate $||\bm{A}_i||_1$~\cite{wong2018provable}. The method of random projections samples standard Cauchy random matrix $\bm{R}$ and split $\bm{R}=\left[\bm{R}_0,\bm{R}_2,\cdots,\bm{R}_{i-1}\right]$ and compute:
\begin{equation}
\bm{A}_i \bm{R} =\bm{W}_i\Lambda_{i-1}\bm{W}_{i-2}\Lambda_{i-3}\cdots\bm{M}_0 \bm{R}_0+\bm{W}_i\Lambda_{i-1}\bm{W}_{i-2}\bm{M}_{i-3}\bm{R}_{i-3}+\bm{W}_i\bm{M}_{i-1}\bm{R}_{i-1}
\end{equation}
Finally, we estimates $||\bm{A}_i||_1 \approx \text{median}(|\bm{A}_i\bm{R}|)$ and compute $\l_i,\u_i$.

Based on these calculations, we search the adversarial examples in layer $i$ as follows:
\begin{equation}
\bm{e}_n \leftarrow \text{clip} (\bm{e}_n + \alpha \bm{A}_i^T \nabla_{\bm{x}_n\prime} \mathcal{ADV}(\bm{x}_n\prime,y_n),-1,1 )
\end{equation}
\begin{equation}
\bm{x}_n\prime \leftarrow \bm{a}_l +\bm{A}_l \bm{e}_n
\end{equation}
where $n$ is the iteration steps, $\text{clip}(\cdot)$ is function which thresholds its argument between -1 and 1.
\subsection{A.3 Certification}
After training completes, we perform certification which is the same as~\cite{balunovic2019adversarial} as follows: for every image, we first try to certify it using only linear relaxations. If this fails, we encode the last layer as MILP and try again. Finally, if this fails we encode the ReLU activation after the last convolution using additional up to 50 binary variables and the rest using the triangle formulation~\cite{ehlers2017formal}. We consider an image to be not certifiable if we fail to certify it using these methods. We always certify the full test set of 10 000 images. The overall verfied pipeline is listed in Algorithm \ref{alg-cert}.

\begin{algorithm} [h]
	\caption{Verfied Pipeline for Provably Robust Certification: $robust(\bm{x}_i)$}  
	\label{alg-cert}
	\begin{algorithmic}
		\STATE {\bfseries Input} {
			$d$-layer network $h_{\bm{\theta}}$, testing sample $\bm{x}_i$, perturbation $\epsilon$.
		}
		\IF[\#\#misclassified]{$h_{\bm{\theta}}(\bm{x}_i) \neq y_i$}   
		\STATE  verfied fails; continue
		\ENDIF
		\IF  {$\bm{x}_i: $ \textit{PGD ATTACK} success }
		\STATE  verfied fails; continue 
		\ENDIF
		
		\IF[\#\#source code: https://github.com/eth-sri/diffai]{$\bm{x}_i: $DIFFAI-v3 verfied}
		\STATE verfied success; continue
		\ENDIF
		
		\STATE  Perform MILP solver to test $\bm{x}_i$ layerwisely as follow. \COMMENT{\#\#source code: https://github.com/eth-sri/colt}
		
		\FOR {$l_i:$ \textit {ALL RELU Layers}}
		\IF{MILP Solver for $\bm{x}_i$ in layer $\#l_i$ fails}
		\STATE \textit{ENCODER RELU layer $\#l_i$} using additional up to 50 binary variables.
		\ENDIF
		\ENDFOR
		
		\IF{MILP Solver for $\bm{x}_i$ success in last layer}
		\STATE verfied success
		\ENDIF
		
 		
\STATE {\bfseries Output} {Verfied or Not Verfied }
	\end{algorithmic}
\end{algorithm}

\subsection{A.4 Evaluation Metrics}
We use four metrics to evaluate our training models: \textit{original accuracy}, \textit{certified robustness}, \textit{verified error}, and \textit{latent robustness}. Original accuracy denotes the fraction of a testing set on which a model is correct. It is the standard accuracy metric used to evaluate any DNN, defended or not. 
Certified robustness denotes the fraction of the testing set on which a certified model predictions
are both correct and certified robust for a given prediction robustness threshold. It has become a standard metric to evaluate models trained with certified defenses~\cite{wong2018provable,raghunathan2018certified}. 
By considering the time consuming for calculating certified robustness during the training process, we define verified error to evaluate the robustness of network in training process. Verified error denotes the fraction of the training set on which the correctly classified examples become to adversarial examples, which is computationally tractable.
Latent robustness denotes the fraction of the testing set on which a model predictions are both correct and robust against latent adversarial attacks on a certain layer. 
Mathematically, the metrics are defined as follows:

\begin{itemize}
	\item \textit{Original Accuracy (ACC):}
	\begin{equation}
	\label{equ:acc}
	\frac{\sum_{i=1}^{n} \mathbb{I} (h_{\bm{\theta}}(\bm{x}_i)=y_i) } {n}
	\end{equation}
	where $\mathbb{I} (h_{\bm{\theta}}(\bm{x}_i)=y_i)$ denotes a function that will return 1 if the prediction on one testing example $\bm{x}_i$ returns the correct label $y_i$, and 0 otherwise.
	\item \textit{Certified Robustness (CR):}
	\begin{equation}
	\label{equ:cr}
	\frac{\sum_{i=1}^{n} \mathbb{I}(robust(\bm{x}_i) )\& (h_{\bm{\theta}}(\bm{x}_i)=y_i))}{n}
	\end{equation}
	where $robust(\bm{x}_i)$ represents the network output is certifiable robust to input $x_i$ (according to Appendix A.3), $\mathbb{I}(\cdot )$ is a indictor function.
	\item \textit{Verified Error (VE):} 
	\begin{equation}
	\label{equ:vr}
	\frac{\sum_{i=1}^{n}\mathbb{I}((h_{\bm{\theta}}(\bm{x}_i^\prime)\neq y_i )\&( h_{\bm{\theta}}(\bm{x}_i)=y_i)) }{\sum_{i=1}^{n} \mathbb{I}(h_{\bm{\theta}}(\bm{x}_i)=y_i)}
	\end{equation}
	where $\bm{x}_i^\prime \in \mathcal{B}_\epsilon(\bm{x}_i)$, $\mathbb{I}(\cdot )$ is a indictor function.
	\item \textit{Latent Robustness (LR):}
	\begin{equation}
	\label{equ:lr}
	\frac{\sum_{i=1}^{n}\mathbb{I}((h_{\bm{\theta}}^{d^{th}}(\bm{x}_i^\prime)= y_i )\&( h_{\bm{\theta}}(\bm{x}_i)=y_i)) }{n}
	\end{equation}
	where $\bm{x}_i^\prime \in \mathcal{B}_\epsilon(\bm{x}_i)$, $\mathbb{I}(\cdot )$ is a indictor function, $h_{\bm{\theta}}^{d^{th}}(\cdot)$ denotes that latent adversarial attack is performed on d-th layer of the network.
	
\end{itemize}

\section{Appendix B. Network Sizes and Training Parameters}

We perform all experiments on a desktop PC using a single GeForce RTX 2080 Ti GPU and 12-core Intel(R) Core(TM) i7-8700 CPU @ 3.20GHz. We implemented training and certification in PyTorch and used Gurobi 9.0 as a MILP solver.
\subsection{B.1 CIFAR-10}
The network architecture used for CIFAR-10 is a 4-layer convolutional network: first 3 layers are convolutional layers with filter sizes 32, 32, 128, kernel sizes 3, 3, 4 and strides 1, 2, 2, respectively. Convolutional layers are followed by a fully connected layer consisting of 250 hidden units. After each layer there is a ReLU activation. Final layer is a fully connected layer with 10 output neurons. The model is trained using SGD with momentum 0.9, the initial learning rate is 0.03 and after the initial 60 epochs we multiply the learning rate by 0.5 every 10 epochs. 
We use 8 steps during the latent adversarial attack and each step size is 0.25 (where perturbations are normalized between -1 and 1).
We perform MAAR in 4 stages (i.e., the first convolutional layer (Stage \#1), the second convolutional layer (Stage \#2), the third convolutional layer (Stage \#3), and the fully connected layer (Stage \#4)), for 200 epochs per stage, for a total of 800 epochs.

\subsection{B.2 MNIST}
For MNIST, we use a 3-layer convolutional network: 2 convolutional layers with kernel sizes 5 and 4, and strides 2 followed by 1 fully connected layer consisting of 100 hidden units. Each of the layers is followed by a ReLU activation function. We use 40 steps for the latent adversarial attack with step size 0.035. We train using Adam~\cite{kingma2014adam} with initial learning rate 0.0001 and after the initial 100 epochs we multiply the learning rate by 0.5 every 10 epochs.
We perform MAAR on MNIST in 4 stages (i.e., the first convolutional layer (Stage \#1), the second convolutional layer (Stage \#2),  and the fully connected layer (Stage \#3)), for 300 epochs per stage, for a total of 900 epochs.

\section{Appendix C. Additional Results}
\subsection{C.1 Sensitivity of Regularization Parameter $\lambda$ }
\begin{figure}[h]
	\centering
	\includegraphics[width=0.5\columnwidth]{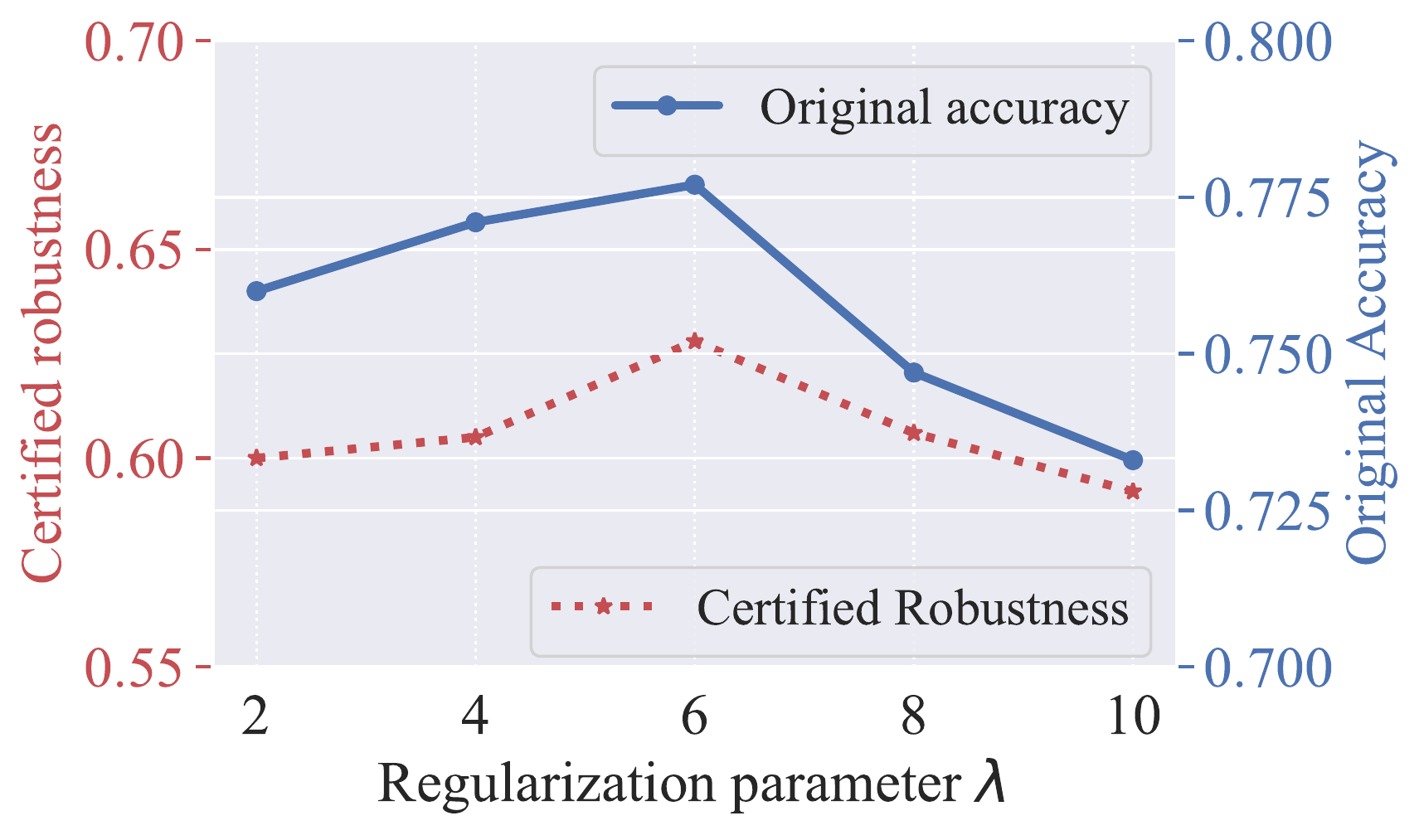} 
	\caption{The network achieves different certified robustness and accuracy  across different choices of regularization parameter $\lambda$.}
	\label{fig:reg}
\end{figure}

We investigate the parameter $\lambda$ with MAAR defined in Equation~(\ref{equ:obj}) which controls the contribution of the regularization term. We present the results in Figure~\ref{fig:reg} for different $\lambda \in \{2, 4, 6, 8, 10\}$. By explicitly setting different impact parameter of misclassified examples, the network achieves good stability and robustness across different choices of $\lambda$. According to the experimental results, we choose $\lambda =6$ for our experiments.

\subsection{C.2 The Effectiveness of MAAR}

\begin{figure}[t]
	\centering                                                          
	\subfigure[Verified Error]{                   
		\begin{minipage}{0.4\columnwidth}
			\centering                                                          
			\includegraphics[width=1\columnwidth]{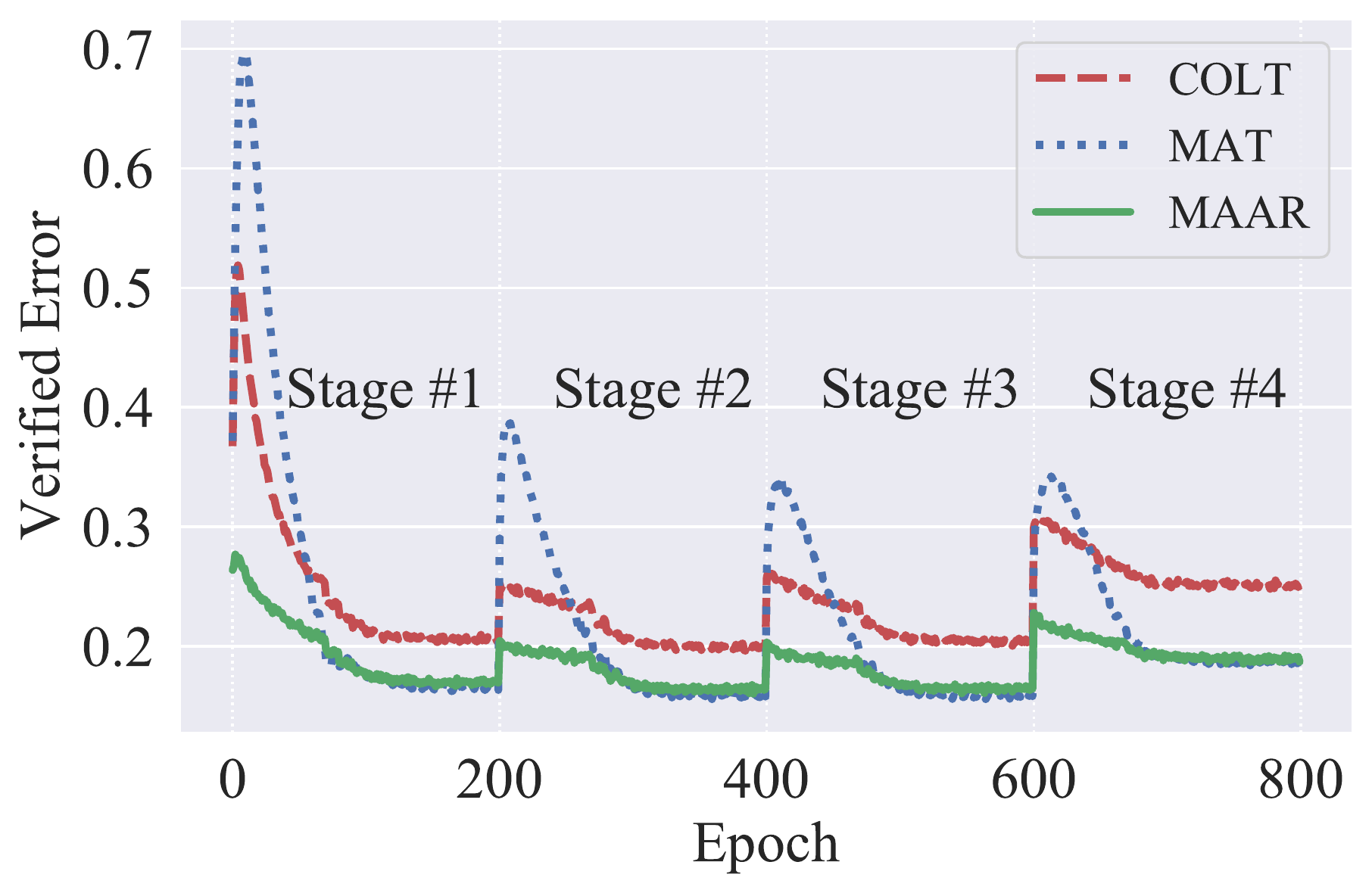}           
	\end{minipage}}
	\subfigure[Original Accuracy]{                  	  
		\begin{minipage}{0.4\columnwidth}
			\centering                                                          
			\includegraphics[width=1\columnwidth]{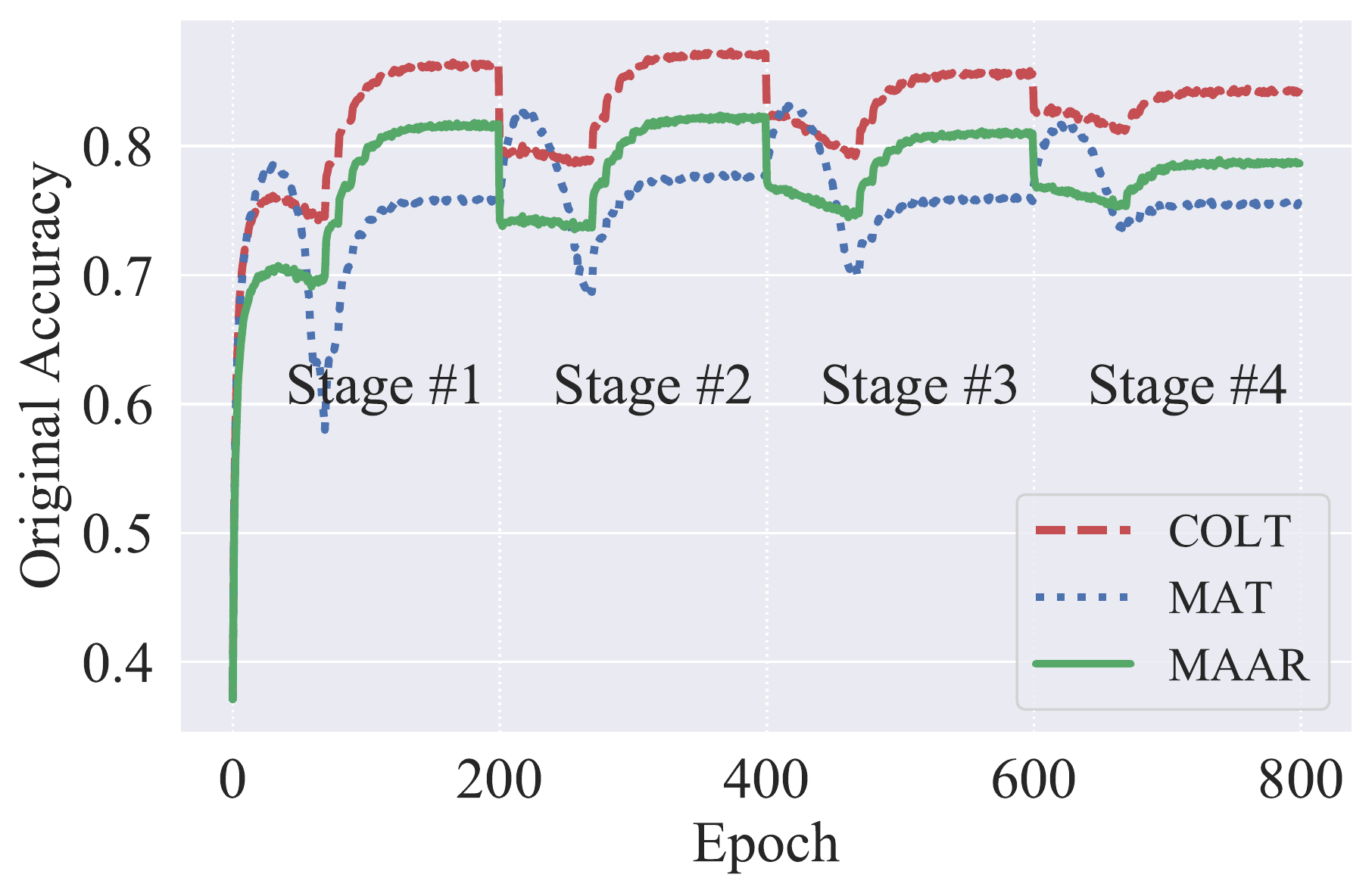}               
	\end{minipage}}
	\caption{Layerwise verified error (a) and layerwise original accuracy (b) of COLT, MAT and our proposed MAAR on different stages. The 4-stage network is trained on the CIFAR-10 dataset with $L_\infty$ perturbation 2/255. }                      
	\label{fig:Cer and Acc}                                             
\end{figure}

\subsubsection{Comparison with COLT and MAT.}
In order to verify the effectiveness of our proposed MAAR, we firstly compare MAAR with COLT and MAT. Note that all experiment settings of these three methods are the same instead of the constraint  on misclassified examples. The verified error and original accuracy evaluated at every epoch during training process has been shown in Figure~\ref{fig:Cer and Acc}.

As shown in Figure~\ref{fig:Cer and Acc}(a),  the verified error of MAAR (green line) decreases more rapidly in comparison with COLT (red line) during each stage, which indicates that MAAR can reduce the proportion of potential adversarial examples in each layer. On the other hand, MAAR maintains the stability of misclassified examples by an additional regularization constraint rather than replacing the training label as MAT, which mitigate the decrease of original accuracy. As shown in Figure~\ref{fig:Cer and Acc}(b), the accuracy of MAAR (green line) is obviously improved when compared with MAT (blue line).

\subsubsection{The Consistent Promotion of MAAR in Layerwise Training Mechanism.}

\begin{table}[h]
	\centering
	\caption{The final certified robustness (CR) and latent robustness (LR) of network trained on MAAR and COLT with the parameters of different stages. LR$^{3^{rd}}$ represents the latent adversarial attack is performed on the 3-rd ReLU layer.}
	\label{tab:diffstages}
	\begin{tabular}{cccc}
		\toprule
		& Method    & CR(\%) & LR$^{3^{rd}}$(\%) \\
		\midrule
		\multirow{2}{*}{Stage \#1}
		& MAAR (Our work) & $\boldsymbol{54.1}$ & $\boldsymbol{58.3}$\\
		& COLT~\shortcite{balunovic2019adversarial}&40.0&47.5\\
		\specialrule{0em}{0pt}{0pt}
		\cmidrule{2-4}
		\specialrule{0em}{0pt}{0pt}
		\multirow{2}{*}{Stage \#2} 
		& MAAR (Our work)& $ \boldsymbol{57.5}$  & $\boldsymbol{60.1}$ \\
		& COLT~\shortcite{balunovic2019adversarial} &48.2 &54.5 \\
		\specialrule{0em}{0pt}{0pt}
		\cmidrule{2-4}
		\multirow{2}{*}{Stage \#3}  
		& MAAR (Our work)& $ \boldsymbol{60.7}$ & $ \boldsymbol{62.0}$  \\
		& COLT~\shortcite{balunovic2019adversarial} & 57.7 &60.8  \\
		\specialrule{0em}{0pt}{0pt}
		\cmidrule{2-4}
		\multirow{2}{*}{Stage \#4} 
		& MAAR (Our work)& $\boldsymbol{62.8}$ & $\boldsymbol{64.7}$  \\
		& COLT~\shortcite{balunovic2019adversarial} & 59.6  & 62.1 \\
		\bottomrule               
	\end{tabular}
\end{table}

In addition, we evaluate the final certified robustness of the network on the checkpoint saved after each stage’s training.
As shown in Table~\ref{tab:diffstages}, we can observe that the final certified robustness of our network has been significantly improved in layerwise training fashion (from 54.1\% on Stage \#1 to 62.8\% on Stage \#4).
Furthermore, the final certified robustness of our proposed MAAR is obviously higher than COLT when evaluated on all stages. 
Meanwhile, we investigate the latent robustness (LR) of the model. Generally, we run latent adversarial attack (i.e., PGD attack with 150 steps and step size of 0.01) on 3-rd ReLU layer with parameters of each stage. Table~\ref{tab:diffstages} indicates the LR of our proposed MAAR improves from 58.3\% on Stage \#1 to 64.7\% on Stage \#4, which is also obviously outperforming COLT on all stages. 
These results demonstrate that MAAR can bring the consistent promotion in layerwise training.

\end{document}